\documentclass[letterpaper, 10 pt, conference]{ieeeconf}
\IEEEoverridecommandlockouts    

\usepackage{times}
\usepackage{amsmath}
\usepackage{amssymb}
\usepackage{graphicx}
\usepackage[table]{xcolor}
\usepackage{tikz}
\usepackage{adjustbox}
\usepackage{bm}
\usepackage{booktabs}
\usepackage{multirow}
\usepackage{siunitx}
\usepackage{pifont}
\usepackage{hyperref}
\usepackage[nocompress]{cite}
\usepackage[font=small]{caption}
\usepackage{xspace}
\usepackage{array}
\usepackage{flushend}

\def\secref#1{Sec.~\ref{#1}}
\def\figref#1{Fig.~\ref{#1}}
\def\tabref#1{Tab.~\ref{#1}}
\def\eqref#1{Eq.~(\ref{#1})}
\makeatletter
\DeclareRobustCommand\onedot{\futurelet\@let@token\@onedot}
\def\@onedot{\ifx\@let@token.\else.\null\fi\xspace}
\def\etal{{et al}\onedot}
\makeatother
\def\etalcite#1{\etal~\cite{#1}}
\newcolumntype{C}[1]{>{\centering\let\newline\\\arraybackslash\hspace{0pt}}m{#1}}

\newcommand{\cmark}{\textcolor{green}{\ding{51}}}
\newcommand{\xmark}{\textcolor{red}{\ding{55}}}
\newcommand{\win}[1]{\cellcolor{gray!25}#1}

\newcommand{\copyrighttext}{\footnotesize \textcopyright{} 2026 IEEE. Personal use of this material is permitted. Permission from IEEE must be obtained for all other uses, in any current or future media, including reprinting/republishing this material for advertising or promotional purposes, creating new collective works, for resale or redistribution to servers or lists, or reuse of any copyrighted component of this work in other works.}
\newcommand{\copyrightnotice}{
\begin{tikzpicture}[remember picture,overlay]
\node[anchor=south,yshift=10pt] at (current page.south) {\fbox{\parbox{\dimexpr0.75\textwidth-\fboxsep-\fboxrule\relax}{\copyrighttext}}};
\end{tikzpicture}
}

\title{\LARGE \bf Robust 6-DoF Object Pose Tracking with Built-In\\
Recovery under Occlusions and Rapid Object Motions}

\author{Balázs Opra$^{1,2}$ \and
  Léo Ghafari$^{1}$ \and
  Thomas Stewart$^{1}$ \and
  Cyrill Stachniss$^{3}$\thanks{\fontsize{6.5pt}{7.2pt}\selectfont (Corresponding author: Balázs Opra.) $^{1}$Woven by Toyota, Inc. $^{2}$University of Bonn, Germany. $^{3}$University of Bonn, Center for Robotics, and the Lamarr Institute for Machine Learning and Artificial Intelligence.
  Email: \texttt{\{balazs.opra, leo.ghafari, tom.stewart\}@woven.toyota},  \texttt{cyrill.stachniss@igg.uni-bonn.de }}}

\IEEEaftertitletext{\vspace{-2mm}}

\begin{document}
\thispagestyle{empty}
\pagestyle{empty}
\maketitle
\AddToHookNext{shipout/foreground}{\copyrightnotice}

\begin{abstract}
\looseness=-1
  Real-time 6-DoF object pose tracking is essential for many robotics applications, and several approaches exist. Yet even today's approaches remain unreliable under temporary full occlusions and rapid object motions. Once
  tracking is lost, most methods struggle to detect the failure and recover automatically,
  often requiring manual re-initialization.
In this paper, we address the problem of robust model-based 6-DoF tracking of unseen objects from RGB-D data, especially in scenarios with occlusion and fast motion.
We propose a novel method that combines efficient learning-based keypoint matching
  with optimization-based alignment and introduces a novel failure detection and recovery module. Our system monitors pose reliability, detects tracking divergence or occlusions, and performs a global re-detection and pose estimation step that robustly verifies recovery candidates before resuming tracking.
Our evaluation on standard tracking benchmarks and on a new dataset of occluded and fast-moving scenes shows that our method matches state-of-the-art accuracy on easy tracking sequences, maintains high tracking speed at 57.6 frames per second, and provides the most robust tracking performance under challenging conditions.
  Thus, we believe that our approach is a relevant step forward in robust 6-DoF object tracking from RGB-D data.

\end{abstract}

\section{Introduction}
\label{sec:intro}

Model-based object tracking, where a 3D model of the target object is available at test time, has seen significant progress in recent years, with
state-of-the-art methods~\cite{wen2024cvpr,stoiber2023iros} starting to saturate classical benchmarks such as YCB-Video~\cite{xiang2018rss}.
These methods enable robotics applications to perform manipulation tasks with higher precision, and to be more reactive to
dynamic events such as grasping failures or in interactions
with humans. Object tracking also serves as the basis for some real-world deployments
of robot learning methods~\cite{huang2023iros,rana2024arxiv}.
However, full or partial occlusions and sudden object motions, such as those that occur when objects are dropped or temporarily go out of frame, are
part of real-world robotics operations and are often not well addressed by object tracking methods.

In this paper, we investigate the problem of accurate and robust model-based object pose tracking under challenging occluded
and dynamic scenarios.
Model-based object tracking is traditionally addressed by establishing correspondences between an
RGB or RGB-D video stream and the object model, and framing the tracking problem as iterative pose
refinement with numerical optimization~\cite{stoiber2023iros}. More recently, learning-based methods became the prominent choice for correspondence estimation~\cite{shaikewitz2024ral}, and in some cases replaced numerical optimization with a neural network~\cite{wen2024cvpr}.
Both optimization and learning-based approaches are capable of highly accurate tracking in slower-moving scenes with partial occlusions,
but tend to fail when confronted with fully losing view of the target object or rapid motions. Most methods are unable to
effectively recover tracking once diverged and require a full re-initialization.

\begin{figure}[t]
  \centering
  \includegraphics[width=0.88\linewidth]{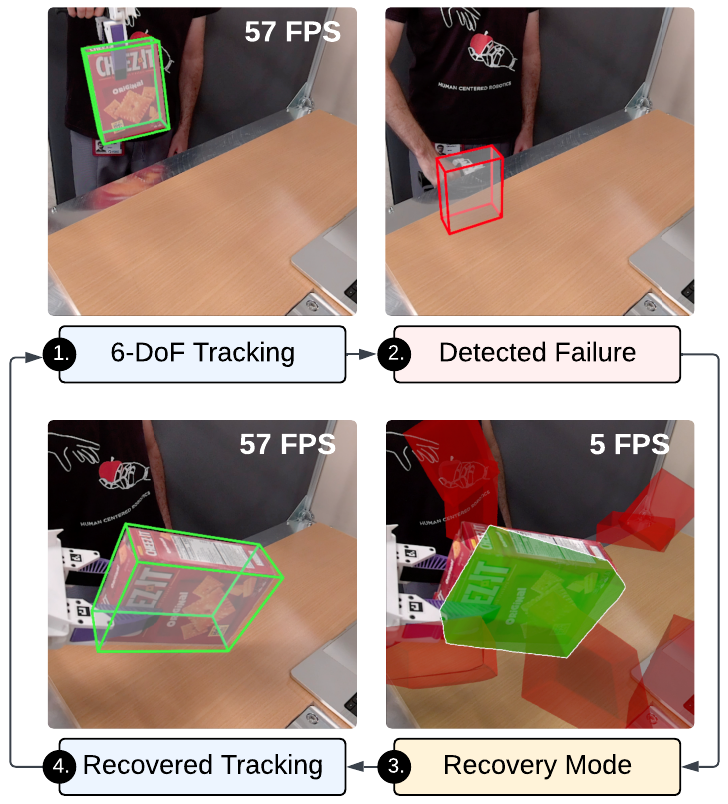}
  \vspace{-2mm}
  \caption{Our 6-DoF object pose tracking system is capable of detecting tracking failure including due to full occlusions, and is able to re-detect the object and resume tracking automatically.}
  \label{fig:motivation}
  \vspace{-2mm}
\end{figure}

\looseness=-2
The main contribution of this paper is a novel accurate model-based RGB-D object pose tracking method that can
detect and handle tracking failures, where the estimated object pose no longer corresponds to the actual object pose due to divergence, rapid object motion, or complete occlusion. Our method builds upon the state-of-the-art ICG{+} tracker~\cite{stoiber2023iros} and is capable of autonomously recovering object tracks once the tracked object becomes visible again, as illustrated in \figref{fig:motivation}.
Our system maintains high tracking speed and recovers from tracking failure with low latency, while
matching the accuracy of leading methods on standard benchmarks.
We make three key claims:
Our approach is able to
(i) reliably detect and recover from tracking failure due to divergence or full occlusions;
(ii) match the state-of-the-art tracking accuracy on general and robotics-focused datasets;
(iii) maintain high tracking speed at 57.6 frames per second and recover from tracking failure with low latency.
These claims are backed up by the paper and our experimental evaluation.

\section{Related Work}
\label{sec:related}

The classical paradigm for tracking a known 3D object model is to
continuously optimize its pose to align with sensor data. Region-based
methods like PWP3D~\cite{prisacariu2012ijcv} segment the object's
silhouette and maximize foreground-background separation. Subsequent
works improve robustness via localized color models around contours
\cite{tjaden2017iccv}, incorporating photometric constraints
\cite{zhong2019ijcv}, and efficient sparse updates~\cite{stoiber2022ijcv}.
Region-based methods excel for texture-poor objects, but can be slow and
struggle with axially symmetric objects.

\looseness=-1
With RGB-D input, many trackers use dense geometric alignment, iterative closest point being a common tool for aligning the object's 3D model to depth data~\cite{kehl2017cvpr}. Other methods use probabilistic
alignment using signed distance fields~\cite{ren2017ijcv}. Many works
combine modalities such as region and depth~\cite{kehl2017cvpr,stoiber2022cvpr}.
Texture keypoint matching is another common approach, yielding 2D-3D
correspondences for pose estimation. Before the widespread adoption of deep learning,
methods matched corners~\cite{harris1988avc,shi1993cornel}, or features such as SIFT~\cite{lowe2004ijcv} to compute frame-to-frame pose updates
\cite{klein2007ismar,vacchetti2004tpami,stoiber2023iros}. 

Recent works integrate
learned components or end-to-end models for tracking. Some employ
deep networks to iteratively refine pose estimates~\cite{li2018arxiv},
or combine learned codes with traditional methods such as particle filters~\cite{deng2019arxiv}.
ROFT~\cite{piga2022ral} combines optical-flow-based pose tracking with an unscented Kalman filter and targets real-time tracking under fast object motion.
VIPose~\cite{ge2021vipose} uses visual-inertial cues for fast-motion tracking in a moving-camera setting exploiting IMU data.
Neural methods like se(3)-TrackNet~\cite{wen2020iros} use synthetic data
to train a neural network to directly predict pose changes between frames.
Many learning-based trackers require extensive per-object training which limits their applicability in practice.
Other works aim to track objects without an exact CAD model, relying on
category priors or online reconstructions
\cite{wang2020icra,wen2021iros,wen2023cvpr,shaikewitz2024ral} at the
cost of heavier computation or slight accuracy trade-offs.

\looseness=-1
The latest trend in object tracking is to train or exploit foundation
models for pose estimation.
FoundationPose~\cite{wen2024cvpr} uses
large synthetic datasets to train a model capable of pose estimation
and tracking of unseen objects given a CAD model or even a few
reference RGB-D images. Other methods use pretrained 3D
foundation models~\cite{oquab2024tmlr,leroy2024eccv}
for pose estimation~\cite{ornek2024eccv,deng2025cvpr} and
tracking~\cite{nguyen2025arxiv} using sim2real matching between
pre-rendered and real images of the target object.
DynamicPose~\cite{liang2025dynamicpose} builds on FoundationPose and extends it for fast moving-camera settings with visual-inertial input.

To our knowledge, the only proposed tracking system that explicitly addresses
tracking reliability is RGBTrack~\cite{guo2025arxiv}, which uses an additional segmentation system~\cite{cheng2022eccv} to
detect tracking failure and a Kalman filter to recover from it. In contrast, we implement a self-contained reliability check and recovery without an external supervision system.

In our approach, we assume that a CAD model or a reconstructed mesh is
available for the target object. Our contribution is a robust optimization-based RGB-D tracking system with an integrated automatic failure
detection and re-initialization module that uses single-frame pose estimation to recover from tracking failure.
The system includes intelligent keyframe management and a 3D point-to-point alignment term from depth, both of which improve
accuracy in regular tracking scenarios. It yields a robust tracker
that outperforms state-of-the-art approaches on challenging tracking
scenarios.

\section{Our Approach to Model-Based Pose Tracking}
\label{sec:main}

\begin{figure*}[t]
  \centering
  \includegraphics[width=0.95\textwidth]{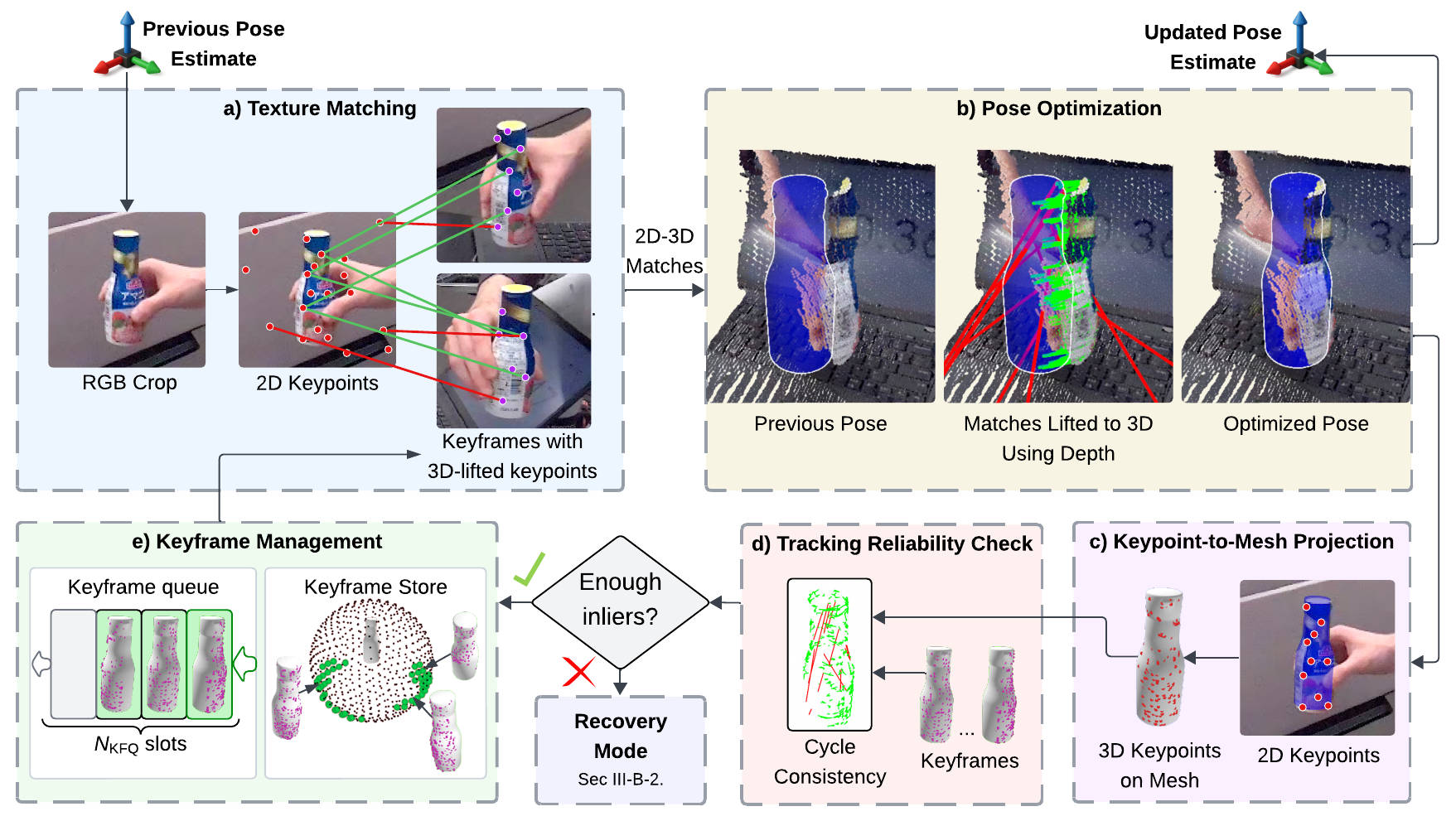}
\caption{Overview of the Texture-Based Tracking System. (a) We perform texture-based keypoint matching between frozen prior frames called keyframes and the current frame. (b) We lift the matches to 3D using the object model and depth data, filter outliers,
  and update the pose estimate via numerical optimization. (c) Using the updated pose,
  we project the image keypoints onto the model, and (d) perform a cycle-consistency-based reliability check. (e) If the check passes, we may store the current frame as a keyframe for future matching. }
  \label{fig:overview}
  \vspace{-3mm}
\end{figure*}

Our system works with one or more RGB-D sensors, and assumes calibrated camera poses, that is,
the homogeneous transformation \( {}_\mathrm{W}\bm{T}_\mathrm{C} \) between the world frame \( \mathrm{W} \) and the camera frame \( \mathrm{C} \) is known. We also
assume calibrated camera intrinsics. Given an initial pose estimate of the target object model frame \( \mathrm{M} \), \( _\mathrm{W}\bm{T}_{\mathrm{M},0} = {}_\mathrm{W}\bm{T}_\mathrm{C} \, {}_\mathrm{C}\bm{T}_{\mathrm{M},0} \), our task is to iteratively update the pose with each incoming RGB-D frame.
To estimate the updated pose for a new frame, we use correspondence data from region, texture, and
depth modalities.

Our approach builds upon ICG{+} proposed by Stoiber~\etalcite{stoiber2023iros}, a multi-modal RGB-D tracker that combines region, depth, and texture correspondences in a single Gauss-Newton optimization.
We keep the region and depth modalities unchanged compared to ICG{+}.
We introduce five components in our proposed method that improve accuracy, especially rotational accuracy, and strengthen robustness under tracking failure: (i) we extend the texture modality with high-throughput \mbox{SuperPoint} TensorRT keypoint extraction and GPU-based keypoint matching to maintain runtime while extracting six times more keypoints; (ii) we add a new 3D distance residual with analytic gradient and Hessian terms to improve tracking accuracy; (iii) we expand keyframe management with an icosahedron-indexed keyframe store to preserve viewpoint diversity; (iv) we integrate a cycle-consistency reliability check to detect divergence and occlusion; and (v) we add a global recovery procedure to handle full tracking failure.
The following subsections detail these components.

\subsection{Optimization-based Tracking}

For completeness, we first briefly restate the optimization-based ICG{+} tracking framework.
The tracker system is composed of three modalities: region, depth, and texture. They are used to establish correspondences
\( \mathcal{D}_{\mathrm{r}}, \mathcal{D}_{\mathrm{d}}, \mathcal{D}_{\mathrm{t}} \)
between the tracked object's body of the previous tracking iteration at \(k-1\)
and the current color and depth data coming from one or more RGB-D cameras.
The exact form of \( \mathcal{D} \) depends on the specific modality.
The depth modality uses model point to depth point correspondences and the region modality
derives correspondences by considering image pixels along correspondence lines which are orthogonal to the model body, as  in Stoiber~\etalcite{stoiber2023iros}. The texture modality derives correspondences from image keypoint matching.

In each tracking iteration, we aim to use the correspondences to find a small pose variation in the model frame \( \bm{\theta}^\mathsf{T} =
  \begin{bmatrix}
    \bm{\theta}_r^\mathsf{T} &
    \bm{\theta}_t^\mathsf{T}
\end{bmatrix} \) which is composed of the rotation vector~\( \bm{\theta}_r \in \mathbb{R}^3 \) and
translation vector~\( \bm{\theta}_t \in \mathbb{R}^3 \), to update the estimated object pose. Specifically, once we found \( \bm{\theta} \) using the latest correspondences, we update the pose estimate as

\begin{equation}
    {}_\mathrm{W}\bm{T}_{\mathrm{M},\,k+1} = {}_\mathrm{W}\bm{T}_{\mathrm{M},\,k} 
    \begin{bmatrix}
        \bm{I} + [\bm{\theta}_r]_{\times} & \bm{\theta}_t \\
        \bm{0} & 1
    \end{bmatrix} \text{,}
\end{equation}
\noindent based on the first-order approximation of the exponential map from the Lie algebra \( \mathrm{se}(3) \) to the Lie group \( \mathrm{SE}(3) \).

We find \( \bm{\theta} \) by maximizing
the posterior probability density function (PDF)

\begin{equation}
  \label{eq:main_pdf}
  p( \bm{\theta} \mid \mathcal{D} ) =
  \prod_{i=1}^{n_\mathrm{r}} p( \bm{\theta} \mid \mathcal{D}_{\mathrm{r}i})
  \prod_{i=1}^{n_\mathrm{d}} p( \bm{\theta} \mid \mathcal{D}_{\mathrm{d}i})
  \prod_{i=1}^{n_\mathrm{t}} p( \bm{\theta} \mid \mathcal{D}_{\mathrm{t}i}),
\end{equation}

\noindent where \(n_\mathrm{r}, n_\mathrm{d}, n_\mathrm{t} \) are the number of region, depth, and texture correspondences, respectively. When multiple RGB-D sensors are available, separate modality instances
can be configured per camera and all contribute correspondences to the
joint optimization, following ICG{+}.

To find a \( \hat{\bm{\theta}} \) that maximizes \eqref{eq:main_pdf}, we use
Gauss-Newton optimization with Tikhonov regularization,

\vspace{-2mm}
\begin{equation}
  \label{eq:main_gauss_newton}
  \hat{\bm{\theta}} = \left(
    -\bm{H} +
    \bm{\Lambda}
  \right)^{-1} \; \bm{g},
  \;
  \bm{\Lambda} =
  \begin{bmatrix}
    \lambda_R \, \bm{I}_{ 3 } & \mathbf{0} \\
    \mathbf{0} & \lambda_t \, \bm{I}_{ 3 }
  \end{bmatrix} \text{ ,}
\end{equation}
where the gradient \( \bm{g} \) and the Hessian \( \bm{H} \) are the first- and second-order
derivatives of the logarithm of the PDF, with the latter approximated using the Gauss–Newton formulation. The parameters \(\lambda_R \) and \( \lambda_t \) are the rotation and translation Tikhonov regularization parameters. Intuitively, the Tikhonov regularization
helps stabilize the optimization in cases where any of the degrees of freedom of the pose
are underconstrained. Below, we describe our texture modality extensions, which constitute our first main contribution.

\subsection{Texture Modality}
\label{sec:texture_modality}

In the texture modality, we use local image feature matching to establish correspondences.
At the beginning of each tracking iteration, we project the tracked object's model
into the color image of each RGB-D sensor using the latest pose estimate. We then crop a square-shaped
region of interest around the projected object model, and extract image features from it.
To take advantage of advances in efficient learning-based image matching, we use the SuperPoint~\cite{detone2017cvpr} feature detector network with an efficient NVIDIA TensorRT~\cite{zhou2022icess} implementation.
This allows us to use a region of interest of size 1024\( \times \)1024
and set the number of keypoints to
1800.

Subsequently, we match the features with those of so-called keyframes saved from previous frames. A keyframe is a frame that is kept static, and we match subsequent frames against it. We create new keyframes as the object moves a certain
amount since the last keyframe, and describe our keyframe management in detail in the next section.
We use nearest-neighbor matching
with \( L2 \) distance as our matching algorithm, and keep matches that pass both, Lowe's ratio test~\cite{lowe2004ijcv}
and the two-view cycle-consistency check~\cite{hartley2004book}. We implement the nearest-neighbor search
with LibTorch~\cite{paszke2019neurips} on the GPU, which allows us to maintain
efficiency despite the increased number of keypoints.
We also tried replacing nearest-neighbor matching with the learned keypoint matcher LightGlue~\cite{lindenberger2023iccv}, which is
strong for global image matching. In our object tracking setting, we did not observe an improved tracking accuracy but an increase in feature matching time, going from 4\,ms to 43\,ms, because each keyframe-current-frame pair requires a separate network inference. This yields a nearly 11-fold increase in the matching stage runtime, making LightGlue prohibitively slow for our use case.
Each match produces a three- or two-dimensional residual through
either 3D point distance or reprojection error, and contributes
to the gradient and Hessian; further details are given in \secref{sec:pose_probability_density}.

\subsubsection{Keyframe management}
Following Stoiber~\etalcite{stoiber2023iros}, we create a new keyframe during
tracking if the orientation of
the object
has changed by more than \( 10^\circ \) since the last keyframe. We maintain a queue of
keyframes of size \( N_\text{KFQ} \), dropping the oldest keyframe if the queue is full.

\textbf{Keyframe creation.} To create a keyframe, we use the latest object pose estimate in the camera frame, \( {}_{\mathrm{C}}\bm{T}_{\mathrm{M}}  = {}_{\mathrm{W}}\bm{T}_{\mathrm{C}}^{-1} \, {}_\mathrm{W}\bm{T}_\mathrm{M} \), to project the detected texture
keypoints onto the surface of the model. Specifically, we transform the model into the camera frame using \( {}_{\mathrm{C}}\bm{T}_{\mathrm{M}} \),
and for each keypoint location we find the intersection of the camera ray with the transformed model using the known intrinsic parameters.
The keyframe stores these intersection points defined in the model frame,
\( {}_\mathrm{M}\bm{X} \), together with their visual feature descriptor vectors.

Note that not all detected keypoints from the region of interest fall onto the surface of the model,
as some may correspond to the background or lie on an occluded part of the object. As in ICG{+}~\cite{stoiber2023iros}, we use the depth image
to filter out occluded keypoints, and only keep those that fall on the unoccluded surface of the object model. In each tracking iteration we match all keyframes in the keyframe queue with the current frame's region of interest.

\textbf{Keyframe store.} Only using a queue of
the most recent keyframes is insufficient for robust
tracking under fast object motions and for recovery
from tracking failure, as previously observed
informative views of the object go out of memory as the object rotates.
Thus, we also maintain a buffer called keyframe store
where we permanently store keyframes from already observed view directions. 
The view direction slots of this buffer are represented by the vertices
of a subdivided icosahedron surrounding the target object, as shown in 
\figref{fig:overview}(e). At the end of each tracking iteration, we insert the current frame into the slot with the closest view direction to the current object pose if (a) the slot is empty, or (b) if the current frame has more keypoints than the stored keyframe.
The keyframe store allows the system to maintain a persistent map of the object's appearance from
all previously observed viewpoints, regardless of how much the object has moved since. This is particularly useful for
fast-moving objects, as well as for redetecting an object after tracking failure.

Thus, in addition to the frames in the keyframe queue, we also select up to \( N_\text{KFS} \) keyframes from the keyframe store for matching during tracking.
These are selected from a \( 30^\circ \) view cone around the current view direction of the object,
while maintaining a minimum distance of \( 5^\circ \) from all other selected keyframes.

\subsubsection{Tracking reliability check}

\begin{figure}
  \vspace{1mm}
  \centering
  \includegraphics[width=\linewidth]{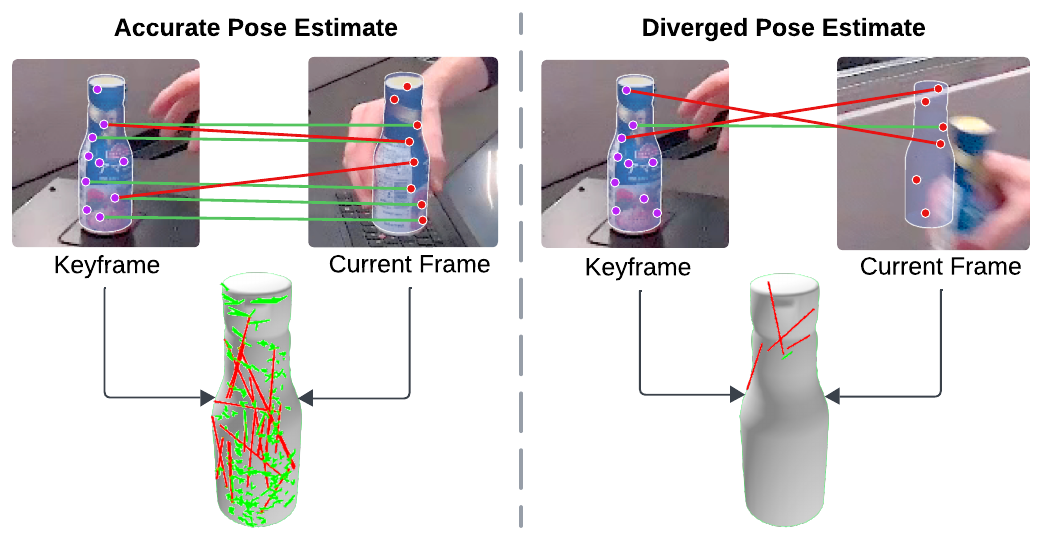}
\caption{Tracking reliability check. The keyframe-current frame matches are projected into the model frame using
  the current pose estimate. Left: inlier (green) and outlier (red)
  matches when the current pose is accurate. Right: very few
  matches and inliers when the pose estimate diverged.}
  \label{fig:reliability_check}
  \vspace{-3mm}
\end{figure}

Since keyframes are created by projecting 2D image keypoints onto the model body using the current pose estimate,
an inaccurate estimate will lead to projected keypoints that do not correspond to their real-world locations on
the model.
To avoid creating inaccurate keyframes and to detect tracking failure, we perform a reliability check using the matches established
between the current frame's region of interest and all active keyframes.

Specifically, at the end of a tracking iteration
we select all matched, unoccluded keypoints
in the current frame which, when projected onto the body mesh using the updated pose estimate,
lie on the surface of the body mesh. We then perform a cycle-consistency check for all such
keypoints,
noting that if (a) the image keypoints matches, and (b) the prior keyframe's pose estimate, and (c) the current pose estimate are all correct, then the
keypoints from the current frame should lie at the same 3D position in the model coordinate system, \( {}_\mathrm{M}\bm{X}^\text{current} \), as their matching keypoints from the keyframes, \( {}_\mathrm{M}\bm{X}^\text{keyframe} \).
Concretely, we deem a keypoint match an inlier if we have the following:
\begin{equation}
  \left\lVert {}_\mathrm{M}\bm{X}^\text{keyframe} - {}_\mathrm{M}\bm{X}^\text{current} \right\rVert _2 <
  \epsilon_\mathrm{cc}, \, \epsilon_\mathrm{cc} \in \mathbb{R}^+ \text{.}
\end{equation}
Refer to \figref{fig:reliability_check} for an illustration.

If there is at least one keyframe with at least \( N_\text{inliers} \) inlier matches, we consider the
current pose estimate acceptable and allow a keyframe to be created from the current frame. An appropriate value for \( N_\text{inliers} \) depends on the number of keypoints we detect. We found \( N_\text{inliers}~= 10\) to work well in all our experiments. If there is more than one configured camera, we perform this check for all cameras individually and allow or disallow the creation of keyframes
per camera.
If none of the cameras
pass the reliability check, we consider the tracking to have failed and explicitly begin the
tracking recovery procedure, described in \secref{sec:failure_recovery}.

\subsubsection{Pose probability density function}
\label{sec:pose_probability_density}

We use two PDFs of pose variation based on different residuals: a 3D distance residual when depth is available and a 2D reprojection residual otherwise.
If the depth image contains a valid depth measurement at a matched keypoint location,
we reconstruct the 3D point \( {}_\mathrm{C}\bm{P} \) from it in the camera frame and
move this point into the model frame using the current pose estimate and the pose variation vector:
\begin{align}
  {}_\mathrm{M}\bm{P}(\bm{\theta}) &= {}_\mathrm{M}\bm{T}(-\bm{\theta}) \, {}_\mathrm{M}\bm{T}_\mathrm{C} \, {}_\mathrm{C}\bm{P} \\
  \text{with } {}_\mathrm{M}\bm{T}(-\bm{\theta}) &= \begin{bmatrix}
        \bm{I} - [\bm{\theta}_r]_{\times} & -\bm{\theta}_t \\
        \bm{0} & 1
    \end{bmatrix}
  \text{.}
\end{align}
We use the 3D distance between \( {}_\mathrm{M}\bm{P}(\bm{\theta}) \) and the matching 3D model point of the keyframe, \( {}_\mathrm{M}\bm{X} \),  as the residual.
We assume a normal distribution for the residuals and formulate the PDF for \( n \)
independent point pairs as
\vspace{-2mm}
\begin{align}
  \label{eq:pdf_texture_3d}
  p( \bm{\theta} | \mathcal{D}_{\mathrm{t}i} ) &\propto \prod_{i=1}^{n} \exp
  \left( -\frac{1}{2 \sigma^2 \mathrm{d}^2} \rho_\text{tuk}
  \left( r_i \right) \right) \\
  \text{with } r_i &= \left\lVert {}_\mathrm{M}\bm{X}_i - {}_\mathrm{M}\bm{P}_i (\bm{\theta}) \right\rVert _2 \text{.}
\end{align}
We scale the residuals by the inverse
of the square of \( \mathrm{d} \), the depth of the correspondence point, which is proportional to the uncertainty of the depth measurement.
Furthermore, the user-defined variance \( \sigma^2 \) is used to
balance the influence of the texture modality with respect to
the other modalities.

To reduce the impact of outliers, we use Tukey's biweight loss
\begin{equation}
  \rho_\text{tuk}(x)  =
  \begin{cases}
    \frac{c^2}{6} ( 1 - ( 1 - ( \frac{x}{c} )^2 )^3 ) & |x| \leq c \\
    \frac{c^2}{6} & |x| > c
  \end{cases} \text{,}
  \label{eq:tukey_loss}
\end{equation}
where \( c \) is typically selected based on the expected magnitude of inter-frame motion of object model surface points.

To derive the gradient and Hessian, we first approximate the Tukey loss as an iteratively reweighted least squares formulation,
\(  \rho_\text{tuk}
  \left( r_i \right) \approx w_i r_i^2,
\)
where the weight is calculated as \( w_i = \rho_\text{tuk} \left( r_i \right) / r_i^2 \),
evaluating the residuals at the current pose estimate with \( \bm{\theta} = \mathbf{0} \). With
this, we can compute the gradient and Hessian contributions from each correspondence using the Gauss-Newton approximation as
\vspace{-1mm}
\begin{align}
  \bm{g}_\mathrm{t} &= -\frac{w_i}{\sigma^2 \mathrm{d}^2}
  ({}_\mathrm{M}\bm{X} - {}_\mathrm{M}\bm{P})^\mathsf{T}
  \begin{bmatrix}
    -[{}_\mathrm{M}\bm{P}]_\times & \bm{I}_3
  \end{bmatrix} \\
  \bm{H}_\mathrm{t} &\approx -\frac{w_i}{\sigma^2 \mathrm{d}^2}
  \begin{bmatrix}
    [{}_\mathrm{M}\bm{P}]_\times \\ \bm{I}_3
  \end{bmatrix}
  \begin{bmatrix}
    -[{}_\mathrm{M}\bm{P}]_\times & \bm{I}_3
  \end{bmatrix} \text{.}
\end{align}

Alternatively, we could vary the fixed model point
\( {}_\mathrm{M}\bm{X} \). This linearization would preserve the exact
point-to-point distance and first-order gradient, while replacing
\( {}_\mathrm{M}\bm{P} \) by \( {}_\mathrm{M}\bm{X} \) in the
Gauss-Newton Hessian, where we could then precompute the unweighted
\([{}_\mathrm{M}\bm{X}]_\times\) factors. In our
ablation on YCBInEOAT~\cite{wen2020iros}, this reduced the
ADD and ADD-S AUC scores~\cite{hinterstoisser2012accv} by \(0.49\) and
\(0.07\) points, so we use the
depth-point linearization above.

Lifting the 2D-3D correspondences to 3D-3D by using the depth information enables us to
filter outlier matches more effectively, as spurious matches between the target object and a nearby background pixel may
have a small reprojection error in the image space, but a large 3D distance error. Although the Tukey loss
helps reduce the influence of such outliers, we find that discarding all matches with a 3D distance residual
larger than a threshold \( \epsilon_\mathrm{3D} > c \) improves accuracy even further. \figref{fig:overview}(b) shows such outlier matches in red. A constant value of \( \epsilon_\mathrm{3D} = 2 c \)
works well in all our experiments.

When there is no valid depth measurement at the keypoint location, we fall back to using
the reprojection error as described by Stoiber~\etalcite{stoiber2023iros}. For an overview of the texture-based tracking system, refer to \figref{fig:overview}.

\subsection{Failure Recovery}
\label{sec:failure_recovery}

\begin{figure*}[t]
  \centering
  \vspace{2mm}
  \includegraphics[width=0.92\textwidth]{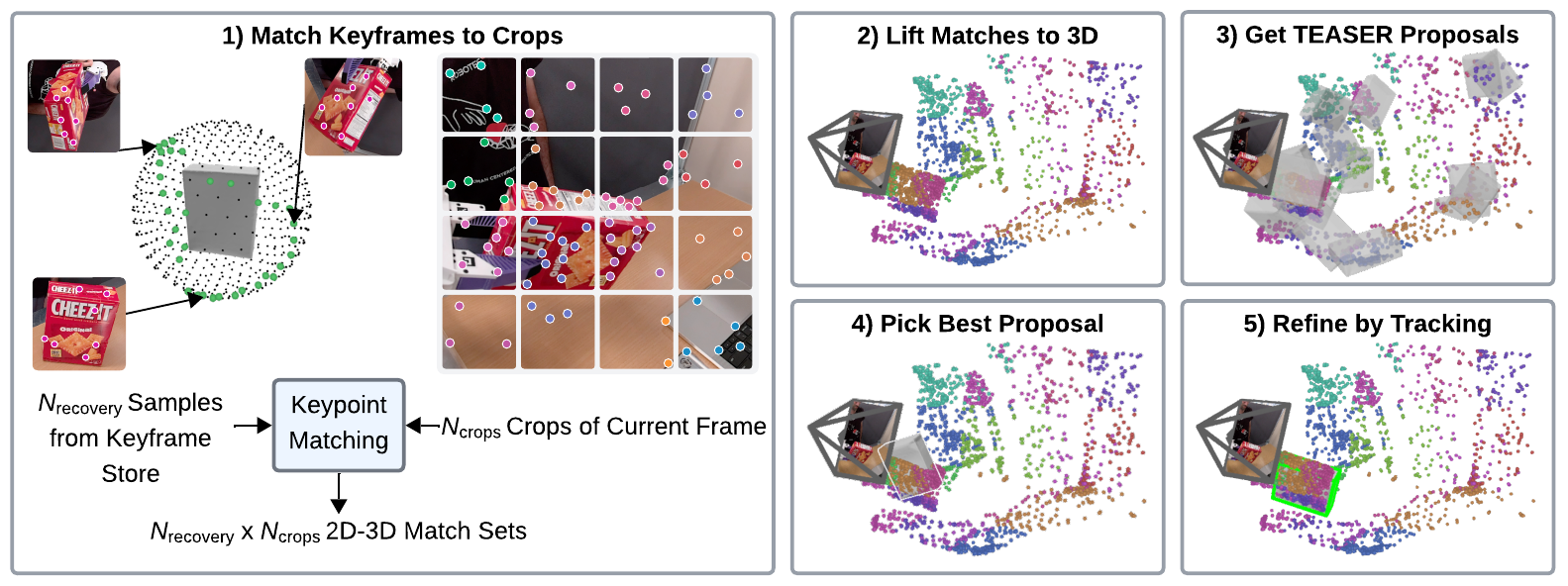}
  \vspace{-2mm}
  \caption{Failure recovery procedure: 1) we match well-selected keyframe
  store keyframes with all crops of the current image, 2) these matches
  are lifted to 3D using depth data, and 3) the TEASER point cloud registration
  method computes pose proposals for each pair. 4) We choose the proposal with the most TEASER inliers. 5) We run one full tracking iteration as described in Sec.~III-A to refine the pose, and then apply the tracking reliability check from Sec.~III-B.}
  \label{fig:recovery_mode}
  \vspace{-3.5mm}
\end{figure*}

Our recovery module is our second main contribution and is integrated into 
the control flow of the tracker: if the reliability check
(\secref{sec:texture_modality}) detects divergence or occlusion, the
system enters recovery mode. Recovery mode performs exhaustive
crop-to-keyframe matching, lifts matches to 3D using depth, solves
robust point cloud registration to obtain candidate object poses, and
resumes tracking only after verifying the candidate pose using the
reliability check. This design avoids any external error detection.

Specifically, in recovery mode, we do not assume any pose prior after a tracking failure, as the object may have moved
rapidly or may have been temporarily occluded. Thus, instead of focusing only on the
region of interest around the last pose estimate, we divide the entire color image
into minimally overlapping crops sized \( 0.25 \) times the longer side of the image, to be used for matching. We have found that this crop size works well for most camera configurations used in robotic manipulation, and produces \( N_\text{crops} = 16 \) crops for a 1280\( \times \)1024 pixel image.
We similarly do not use any pose prior for selecting keyframes, and
instead use a strategy aimed at maximizing the view direction coverage of the object.
In particular, we use farthest-point sampling to select
\( N_\text{recovery} \) keyframes from the keyframe store
whose view direction vectors are maximally distant from each other.

We extract SuperPoint features from the crops and match each crop with
all selected keyframes using the same matching strategy as described in
\secref{sec:texture_modality}. This gives us a set of 2D-3D crop-to-model point
correspondences for each crop. We then use the depth information to reconstruct the 3D points
from the crops to create 3D-3D correspondences between the depth point cloud and the
object model. We solve the resulting point cloud registration problems using TEASER++
\cite{yang2020tro}, which is a fast and robust registration method, to obtain
\( N_\text{crops} \times N_\text{recovery} \) pose estimates. The best candidate
pose is selected as the one with the highest number of inlier keypoints retrieved from the registration solution.

Since the recovery pose proposals are coarse, we re-initialize tracking from the best proposal and run one full tracking iteration as described in Sec.~III-A, which acts as a
pose refinement step. We then perform the tracking reliability check described in Sec.~III-B, and exit recovery mode and continue tracking from the recovered pose estimate if it passes.
If the reliability check fails, we discard the
result of the recovery attempt and continue repeating the recovery procedure for each
new frame until the tracking reliability check passes. The reliability check helps
avoid spuriously accepting incorrect recovery pose proposals. Refer to \figref{fig:recovery_mode} for an illustration of the 
recovery procedure.

During recovery, instead of matching only a single image crop around the last pose estimate, we match multiple crops of the current frame with keyframes and additionally
solve a point cloud registration problem for each crop. This makes computing the recovery pose estimate
more expensive than a regular tracking iteration. Specifically, setting \( N_\text{crops} = 16 \) and \( N_\text{recovery} = 20 \), as we used in all our experiments, we found that it takes about
210\,\si{ms} on average to compute the recovery pose estimate, which is around 12 times longer than a regular tracking iteration.
A natural alternative to our recovery procedure would be to re-initialize tracking with a full object \mbox{segmentation} and pose estimation pipeline, such as CNOS~\cite{nguyen2023iccvws} followed by FoundationPose~\cite{wen2024cvpr} in pose estimation mode. In our setting, this pipeline takes 2.69\,\si{s}, which is more than 10 times longer than our recovery procedure. For a moving object, such a delay makes the returned pose estimate stale by the time it becomes available, which would make using such a pipeline for recovery unreliable.

\section{Experimental Evaluation}
\label{sec:exp}

We present our experiments to support our three key claims: (i) our method reliably detects and recovers from tracking failure due to divergence or full occlusions; (ii) it matches state-of-the-art tracking accuracy on general and robotics-focused datasets; and (iii) it maintains high tracking speed at 57.6 frames per second and recovers from tracking failure with low latency.

\subsection{Experimental Setup}

We evaluate on YCB-Video~\cite{xiang2018rss}, YCBInEOAT~\cite{wen2020iros}, Fast-YCB~\cite{piga2022ral}, and a new RGB-D dataset for rapid motion and full occlusion; its release awaits institutional review. Dataset page: {\scriptsize\href{https://github.com/Bazs/robust-6d-object-pose-tracking}{github.com/Bazs/robust-6d-object-pose-tracking}}. We captured its 6 sequences and 2491 frames at 1280 x 1024 resolution with two overlapping Framos FSM-IMX304 stereo pairs, one for tracking and one for annotation verification. We use learning-based stereo depth~\cite{shankar2022ral}, annotate poses primarily by hand, and initialize a subset of frames with FoundationPose~\cite{wen2024cvpr}. The objects are the \mbox{YCB-V} cracker box and mug plus an axially symmetric textured bottle shown in \figref{fig:overview}. We split the sequences into fast motion-blur cases and occluded cases where the target leaves the camera view.

We compare against PoseRBPF~\cite{deng2019arxiv}, a particle-filter tracker with per-object retraining; ROFT~\cite{piga2022ral}, an optical-flow and Kalman-filter tracker for fast motion; ICG{+}~\cite{stoiber2023iros}, the \mbox{RGB-D} baseline on which our method builds; and FoundationPose~\cite{wen2024cvpr}, a single-view pose estimation and tracking baseline. We report all five methods on the in-house dataset and subsets on YCB-Video, YCBInEOAT, and Fast-YCB\@.

We report ADD and ADD-S area-under-curve metrics~\cite{hinterstoisser2012accv} and measure runtime on a desktop system with an Intel Core i9-14900K CPU and an NVIDIA GeForce RTX 4090 GPU\@.

\subsection{Tracking Failure Recovery}

\looseness=-1
The first experiment supports the claim that our system remains robust under rapid object motion and occlusion and can recover from tracking failure. We evaluate all five methods on the ``fast'' and ``occluded'' sequences of our dataset. To account for the recovery latency of our method, we freeze the output pose for 6 frames after recovery, matching the average recovery latency measured online at 30 FPS\@.

\looseness=-1
Table~\ref{tab:inhouse_accuracy} reports per-object, per-setting, and overall scores, including an ablation with failure detection and recovery disabled. Baseline methods score lower because they lose track during fast motions and occlusions and have limited recovery capability. The recovery-disabled variant of our method achieves the second-best overall ADD score, indicating strong continuous tracking without re-localization. Relative to this ablation, the full system gains 32.77 ADD AUC points and 30.55 ADD-S AUC points, reaching the best overall scores and showing that recovery is also essential under severe tracking failures. Across the six in-house recordings, the full system entered recovery only seven times, indicating that tracking generally proceeds without re-localization. It spent 10.74\% of evaluated frames in recovery, with ratios of 6.58\% for fast sequences and 14.65\% for occluded sequences. Its weakest result is on the textureless, rotationally symmetric \mbox{YCB-V} mug, a difficult case for the texture-focused matching stage, yet it still outperforms all baselines.

\begin{table}[t]
  \vspace{1mm}
  \caption{Tracking accuracy on our dataset for PoseRBPF (PRBPF)~\cite{deng2019arxiv}, ROFT~\cite{piga2022ral}, ICG{+}~\cite{stoiber2023iros}, FoundationPose (FP)~\cite{wen2024cvpr}, Ours without recovery, and Ours. The novel object row indicates whether the method supports unseen object tracking. The best result is bold and the second-best result is underlined.} \centering
  \scriptsize
  \renewcommand{\arraystretch}{.78}
  \setlength{\tabcolsep}{2.8pt}
  \newcommand{\subtabrule}{\cmidrule(l){2-8}}
  \begin{adjustbox}{max width=\columnwidth}
  \begin{tabular}{ll|*{6}{C{0.84cm}}}
    \toprule
    Group & Metric & PRBPF & ROFT & ICG{+} & FP & \shortstack{Ours\\w/o rec.} & Ours \\
    \midrule
    Novel object & & \xmark & \xmark & \cmark & \cmark & \cmark & \cmark \\
    \midrule
    \multirow{2}{*}{bottle}
      & ADD   & 43.69 & 27.74 & 48.58 & 44.99 & \underline{64.24} & \textbf{90.46} \\
      & ADD-S & 47.44 & 32.75 & 56.84 & 46.84 & \underline{64.81} & \textbf{91.09} \\
    \subtabrule
    \multirow{2}{*}{mug}
      & ADD   & \underline{60.57} & 22.35 & 36.29 & 48.19 & 40.32 & \textbf{66.09} \\
      & ADD-S & \underline{69.04} & 32.99 & 55.32 & 48.65 & 51.58 & \textbf{77.33} \\
    \subtabrule
    \multirow{2}{*}{cracker\_box}
      & ADD   & 42.90 & 26.21 & 45.59 & 43.94 & \underline{49.79} & \textbf{92.66} \\
      & ADD-S & \underline{68.57} & 29.04 & 53.06 & 43.94 & 55.96 & \textbf{93.25} \\
    \midrule
    \multirow{2}{*}{Fast}
      & ADD   & \underline{65.46} & 26.70 & 46.71 & 44.89 & 47.60 & \textbf{88.95} \\
      & ADD-S & \underline{74.88} & 31.18 & 64.26 & 46.24 & 53.84 & \textbf{92.05} \\
    \subtabrule
    \multirow{2}{*}{Occluded}
      & ADD   & 32.95 & 24.23 & 40.50 & 46.23 & \underline{53.99} & \textbf{78.70} \\
      & ADD-S & 51.33 & 31.47 & 46.00 & 46.25 & \underline{60.13} & \textbf{83.47} \\
    \midrule
    \multirow{2}{*}{\textbf{All Frames}}
      & ADD   & 48.69 & 25.43 & 43.51 & 45.58 & \underline{50.90} & \textbf{83.67} \\
      & ADD-S & \underline{62.74} & 31.33 & 54.84 & 46.25 & 57.08 & \textbf{87.63} \\
    \bottomrule
  \end{tabular}
  \end{adjustbox}
  \label{tab:inhouse_accuracy}
\end{table}

\subsection{Accurate and Fast Tracking}
\vspace{-2mm}
The second experiment evaluates accuracy and runtime on YCB-Video, YCBInEOAT, and Fast-YCB\@. Tables~\ref{tab:large_accuracy_resource_comparison} and~\ref{tab:fast_ycb_accuracy_resource_comparison} compare against the applicable baselines. The results support our claim that our approach achieves a tracking accuracy similar to the state-of-the-art method FoundationPose while being around 40\% faster on the same hardware. Although ICG{+}~\cite{stoiber2023iros} is faster, our method is more accurate, especially for rotation: it gains 5.09 ADD points on Fast-YCB all frames and 10.86 ADD points on the axially symmetric tomato soup can sequence in YCBInEOAT\@.

\begin{table}[t]
\caption{Tracking accuracy and speed on YCB-Video and \mbox{YCBInEOAT}. The best all-frame result is bold, the second-best is underlined, and per-object winners have a gray highlight.} \centering
  \scriptsize                         \renewcommand{\arraystretch}{.85}   \setlength{\tabcolsep}{4pt}         \begin{adjustbox}{max width=\columnwidth}
    \begin{tabular}{l |
        *{2}{c} |  *{2}{c} |  *{2}{c} |  *{2}{c}}   \toprule
      Approach &
      \multicolumn{2}{c|}{PRBPF} &
      \multicolumn{2}{c|}{ICG\texttt{+}} &
      \multicolumn{2}{c|}{FP} &
      \multicolumn{2}{c}{Ours} \\
      \midrule
      Novel object &
      \multicolumn{2}{c|}{\xmark} &
      \multicolumn{2}{c|}{\cmark} &
      \multicolumn{2}{c|}{\cmark} &
      \multicolumn{2}{c}{\cmark} \\
      Metric &
      {\tiny ADD} & {\tiny ADD-S} & {\tiny ADD} & {\tiny ADD-S} & {\tiny ADD} & {\tiny ADD-S} &
      {\tiny ADD} & {\tiny ADD-S} \\
      \midrule
\multicolumn{9}{@{}l}{\textbf{YCB-Video}}\\
      master\_chef\_can   & 90.5 & 95.1 & 94.7 & 98.0 & 93.6 & 97.0 & \win{96.1} & \win{98.4}\\
      cracker\_box        & 88.2 & 93.0 & 96.8 & \win{98.4} & \win{96.9} & 97.8 & 96.2 & 98.2\\
      sugar\_box          & 92.9 & 95.5 & 96.6 & 98.5 & 96.9 & 98.2 & \win{97.5} & \win{98.9}\\
      tomato\_soup\_can   & 90.0 & 93.8 & 95.2 & 98.1 & \win{96.3} & 98.1 & 95.3 & \win{98.2}\\
      mustard\_bottle     & 91.9 & 96.3 & 97.1 & 98.7 & 97.3 & 98.4 & \win{97.6} & \win{99.0}\\
      tuna\_fish\_can     & 91.1 & 95.3 & 94.1 & 97.1 & \win{96.9} & \win{98.5} & 95.8 & 97.8\\
      pudding\_box        & 85.8 & 92.0 & 80.4 & 90.5 & \win{97.8} & \win{98.5} & 76.3 & 89.4\\
      gelatin\_box        & 96.3 & 97.5 & 96.8 & \win{99.1} & \win{97.7} & 98.5 & 96.2 & \win{99.0}\\
      potted\_meat\_can   & 68.7 & 77.9 & 95.4 & 98.1 & 95.1 & 97.7 & \win{95.6} & \win{98.2}\\
      banana              & 74.2 & 86.9 & 92.8 & 98.2 & \win{96.4} & \win{98.4} & 95.1 & \win{98.4}\\
      pitcher\_base       & 86.8 & 94.2 & 97.2 & 98.9 & 96.7 & 98.0 & \win{97.6} & \win{99.0}\\
      bleach\_cleanser    & 86.0 & 93.0 & 91.6 & 97.0 & \win{95.5} & \win{97.8} & 93.6 & 97.7\\
      bowl                & 25.5 & 94.2 & 84.1 & 97.9 & \win{95.2} & 97.6 & 84.5 & \win{98.0}\\
      mug                 & 90.9 & 97.1 & 95.6 & 98.4 & 95.6 & 97.9 & \win{96.0} & \win{98.6}\\
      power\_drill        & 93.9 & 96.1 & 96.8 & 98.7 & \win{96.9} & 98.2 & 96.8 & \win{98.6}\\
      wood\_block         & 20.1 & 89.1 & 91.7 & 96.7 & 93.2 & 97.0 & \win{94.8} & \win{97.7}\\
      scissors            & 76.1 & 85.6 & \win{94.9} & \win{97.6} & 94.8 & 97.5 & 93.6 & 97.1\\
      large\_marker       & 92.0 & 97.1 & 94.0 & 98.1 & \win{96.9} & \win{98.6} & 92.5 & 97.8\\
      large\_clamp        & 48.5 & 94.8 & 93.8 & 97.8 & 93.6 & 97.3 & \win{94.7} & \win{98.0}\\
      extra\_large\_clamp & 40.3 & 90.1 & 91.8 & 97.0 & \win{94.4} & \win{97.5} & \win{94.4} & \win{97.9}\\
      foam\_brick         & 81.1 & 95.7 & 94.0 & 97.9 & \win{97.9} & \win{98.6} & 95.0 & 98.0\\
      \midrule
      \textbf{All frames} & 80.8 & 93.3 & 94.3 & \underline{97.9} & \textbf{96.0} & \underline{97.9} & \underline{95.0} & \textbf{98.1} \\
      \midrule
\multicolumn{9}{@{}l}{\textbf{YCBInEOAT}}\\
      cracker\_box        & 81.72 & 89.76 & 91.58 & \win{95.63} & 91.32 & 95.10 & \win{92.25} & 95.47 \\
      bleach\_cleanser    & 70.79 & 86.84 & 92.06 & 96.07 & 91.45 & 95.96 & \win{92.32} & \win{96.17} \\
      sugar\_box          & 88.64 & 93.44 & 93.32 & 96.49 & \win{94.14} & 96.67 & 94.06 & \win{96.77} \\
      tomato\_soup\_can   & 91.18 & 96.16 & 79.83 & 95.71 & \win{91.71} & 96.58 & 90.69 & \win{96.60} \\
      mustard\_bottle     & 92.20 & 96.47 & 95.53 & 97.81 & \win{96.34} & 97.89 & 95.85 & \win{97.92} \\
      \midrule
      \textbf{All frames} & 85.54 & 92.68 & 90.79 & 96.35 & \underline{93.09} & \underline{96.42} & \textbf{93.14} & \textbf{96.57} \\
      \midrule
FPS [Hz] &
      \multicolumn{2}{c}{21.7} &
      \multicolumn{2}{|c}{\textbf{111.0}} &
      \multicolumn{2}{|c}{41.3} &
      \multicolumn{2}{|c}{\underline{57.6}} \\
      \bottomrule
    \end{tabular}
  \end{adjustbox}
  \label{tab:large_accuracy_resource_comparison}
\end{table}

\begin{table}[t]
  \vspace{1mm}
  \caption{Tracking accuracy on Fast-YCB\@. The best all-frame result is bold, the second-best is underlined, and per-object winners have a gray highlight.} \vspace{-1mm}
  \centering
  \scriptsize
  \renewcommand{\arraystretch}{.85}
  \setlength{\tabcolsep}{4pt}
  \begin{adjustbox}{max width=\columnwidth}
    \begin{tabular}{l |
        *{2}{c} |  *{2}{c} |  *{2}{c} |  *{2}{c}}   \toprule
      \multicolumn{9}{@{}l}{\textbf{Fast-YCB}}\\
      \midrule
      Approach &
      \multicolumn{2}{c|}{ROFT} &
      \multicolumn{2}{c|}{ICG\texttt{+}} &
      \multicolumn{2}{c|}{FP} &
      \multicolumn{2}{c}{Ours} \\
      Metric & {\tiny ADD} & {\tiny ADD-S} & {\tiny ADD} & {\tiny ADD-S} & {\tiny ADD} & {\tiny ADD-S} & {\tiny ADD} & {\tiny ADD-S} \\ \midrule
      cracker\_box      & 78.50 & 88.05 & 98.69 & 98.75 & 98.39 & 98.53 & \win{99.39} & \win{99.40} \\
      sugar\_box        & 81.15 & 89.87 & 99.01 & 99.11 & 98.94 & 99.11 & \win{99.32} & \win{99.39} \\
      tomato\_soup\_can & 79.00 & 89.74 & 81.25 & 98.46 & \win{99.44} & \win{99.49} & 96.58 & 99.40 \\
      mustard\_bottle   & 73.10 & 86.78 & 98.91 & 99.14 & 98.59 & 98.79 & \win{99.10} & \win{99.38} \\
      gelatin\_box      & 74.26 & 85.49 & 90.58 & 96.04 & \win{99.47} & \win{99.49} & 99.39 & 99.47 \\
      potted\_meat\_can & 73.87 & 87.58 & 94.26 & 98.15 & \win{99.46} & \win{99.49} & 99.43 & 99.48 \\
      \midrule
      \textbf{All frames} & 76.59 & 87.88 & 93.78 & 98.27 & \textbf{99.05} & \underline{99.15} & \underline{98.87} & \textbf{99.42} \\
      \bottomrule
    \end{tabular}
  \end{adjustbox}
  \label{tab:fast_ycb_accuracy_resource_comparison}
\end{table}

\subsection{Ablation Study}
\label{sec:ablation}
The third experiment validates the effectiveness of our proposed 3D texture matching PDF, the use of the SuperPoint keypoint detector, and the keyframe store to improve tracking accuracy. \tabref{tab:abl_add_adds} starts from 300 SIFT keypoints and one keyframe (\(N_\text{KFQ} = 1\)), then adds the 3D residual, SuperPoint with 1800 keypoints, \( N_\text{KFQ} = 9 \), and finally the keyframe store with \( N_\text{KFS} = 10 \). The extensions increase the accuracy in general, with a relative 2.6\% gain in the ADD score, highlighting the improvement in rotation tracking accuracy, which is important for robotic manipulation tasks.

\begin{table}[t]
  \caption{Ablation of tracking settings on YCBInEOAT.}
  \vspace{-2mm}
  \centering
\begin{tabular}{l|c c}
    \toprule
    \textbf{Settings} & {\textbf{ADD}} & {\textbf{ADD-S}} \\
    \midrule
    SIFT, 1 KF, reprojection error residual     & 90.79 & 96.35 \\
    + 3D distance residual                      & 91.53 & 96.36 \\
    + SuperPoint instead of SIFT            & 92.60 & 96.50 \\
    + Increase to 9 keyframes                          & 93.00 & 96.48 \\
    + Enable KF store with 10 keyframes     & \textbf{93.14} & \textbf{96.57} \\
    \bottomrule
  \end{tabular}
  \label{tab:abl_add_adds}
\end{table}

\vspace{-1mm}
\subsection{Limitations}
Since our recovery procedure is based on texture matching with previously seen views of the target object, its effectiveness is limited by the number of view directions observed since
the tracking began. This limitation could be mitigated by
pre-filling the keyframe store with pre-rendered views of the object if a textured model is available, or with real images taken from a variety of orientations.
Moreover, when multiple same-kind instances are present, recovery after full tracking failure can reinitialize on the wrong instance and cause an identity switch.

\section{Conclusion}
\label{sec:conclusion}

\looseness=-1
In this paper, we presented a novel approach to model-based unseen object pose tracking using RGB-D data. Our proposed tracking reliability check, novel keyframe store, and failure recovery procedure allow the system to detect and recover from tracking failure caused by fast object motions or full occlusions, making it the most robust method among the strong and diverse baselines we evaluated in these challenging scenarios. By combining efficient deep learning-based keypoint detection with robust texture matching alignment that uses color and depth information holistically, the method improves tracking accuracy.
Our experiments demonstrate the accuracy and efficiency of our approach on standard benchmarks and highlight its robustness to fast object motions and full occlusions.

\clearpage
\bibliographystyle{plain_abbrv}

\bibliography{glorified,new}
\end{document}